\documentclass[10pt,twocolumn,letterpaper]{article}

\usepackage{cvpr}              % To produce the CAMERA-READY version
\newcommand{\minisection}[1]{\noindent{\textbf{#1}.}}
\usepackage{graphicx}
\usepackage{mwe}
\usepackage{textcomp}
\usepackage[accsupp]{axessibility}

\definecolor{cvprblue}{rgb}{0.21,0.49,0.74}
\usepackage[accsupp]{axessibility}
\usepackage{array}
\definecolor{lightgray}{gray}{0.9}
\definecolor{cvprblue}{rgb}{0.21,0.49,0.74}
\usepackage[pagebackref,breaklinks,colorlinks,citecolor=cvprblue]{hyperref}

\usepackage{colortbl}
\usepackage{xcolor}
\definecolor{lightgray}{gray}{0.9}
\usepackage{booktabs}
\usepackage{color, soul}

\definecolor{lightgray}{gray}{0.9}
\definecolor{lightblue}{rgb}{0.93,0.95,1.0}
\definecolor{darkgreen}{rgb}{0.0,0.6,0.0}
\definecolor{darkblue}{rgb}{0.0,0.0,0.5}
\definecolor{pinegreen}{rgb}{0.0, 0.47, 0.44}
\definecolor{deepmagenta}{rgb}{0.8, 0.0, 0.8}
\definecolor{amber}{rgb}{1.0, 0.49, 0.0}

\newcommand{\ignorebig}[1]{}

\newlength\savewidth

\newcommand{\gcol}[1]{{\bf \fontsize{6.5}{42}\selectfont \color{citecolor!80}~(#1)}}

\definecolor{citecolor}{RGB}{34,139,34}
\definecolor{lightred}{RGB}{241,140,142}
\definecolor{amber(sae/ece)}{rgb}{1.0, 0.49, 0.0}
\definecolor{battleshipgrey}{rgb}{0.52, 0.52, 0.51}
\definecolor{cadmiumorange}{rgb}{0.93, 0.53, 0.18}
\definecolor{applegreen}{rgb}{0.55, 0.71, 0.0}
\definecolor{cadmiumgreen}{rgb}{0.0, 0.42, 0.24}
\definecolor{forestgreen}{rgb}{0.13, 0.55, 0.13}
\definecolor{red}{rgb}{0.89, 0.0, 0.13}

\def\paperID{31640} % *** Enter the Paper ID here 
\def\confName{CVPR}
\def\confYear{2026}

\title{Visual Funnel: Resolving Contextual Blindness\\in Multimodal Large Language Models}

\author{
    Woojun Jung \quad 
    Jaehoon Go \quad 
    Mingyu Jeon \quad
    Sunjae Yoon \quad
    Junyeong Kim\thanks{Corresponding author.}\\
    Department of AI, Chung-Ang University \\
    {\tt\small \{svvma91, gkdwngo, smart2557, sunjaeyoon, junyeongkim\}@cau.ac.kr}
}

\begin{document}
\maketitle
\begin{abstract}
Multimodal Large Language Models (MLLMs) demonstrate impressive reasoning capabilities, but often fail to perceive fine-grained visual details, limiting their applicability in precision-demanding tasks. 
While methods that crop salient regions of an image offer a partial solution, we identify a critical limitation they introduce: ``Contextual Blindness.'' 
This failure occurs due to structural disconnect between high-fidelity details (from the crop) and the broader global context (from the original image), even when all necessary visual information is present.
We argue that this limitation stems not from a lack of information `Quantity,' but from a lack of `Structural Diversity' in the model's input.
To resolve this, we propose Visual Funnel, a training-free, two-step approach. 
Visual Funnel first performs Contextual Anchoring to identify the region of interest in a single forward pass. 
It then constructs an Entropy-Scaled Portfolio that preserves the hierarchical context---ranging from focal detail to broader surroundings---by dynamically determining crop sizes based on attention entropy and refining crop centers. 
Through extensive experiments, we demonstrate that Visual Funnel significantly outperforms naive single-crop and unstructured multi-crop baselines. 
Our results further validate that simply adding more unstructured crops provides limited or even detrimental benefits, confirming that the hierarchical structure of our portfolio is key to resolving Contextual Blindness.
\end{abstract}    
\vspace{-0.3cm}
\begin{figure}[!t]
    \centering
    \includegraphics[width=1.0\linewidth]{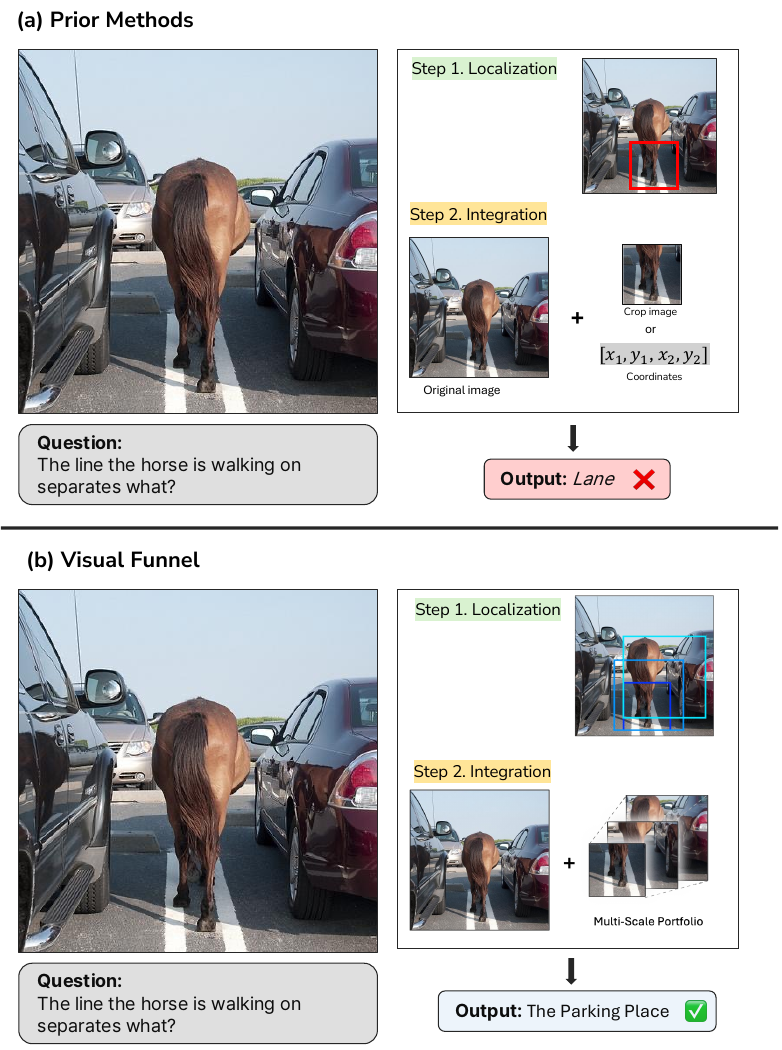}
    \caption{
    \textbf{Illustration of Contextual Blindness and our proposed solution, Visual Funnel.}
    \textbf{(a)} Prior single-crop methods successfully localize the area of interest but perform a naive integration by providing only a tight crop. This isolates the detail from its necessary context, leading to an incorrect answer (e.g., misidentifying a `parking space' as a `lane').
    \textbf{(b)} Our Visual Funnel performs a more sophisticated integration by generating a multi-scale portfolio. This portfolio preserves the hierarchical context, enabling the MLLM to resolve the ambiguity and provide the correct answer.
    }
    \label{fig:teaser}
    \vspace{-0.4cm}
\end{figure}

\begin{figure*}[!t]
  \centering
     \includegraphics[width=1.0\linewidth]{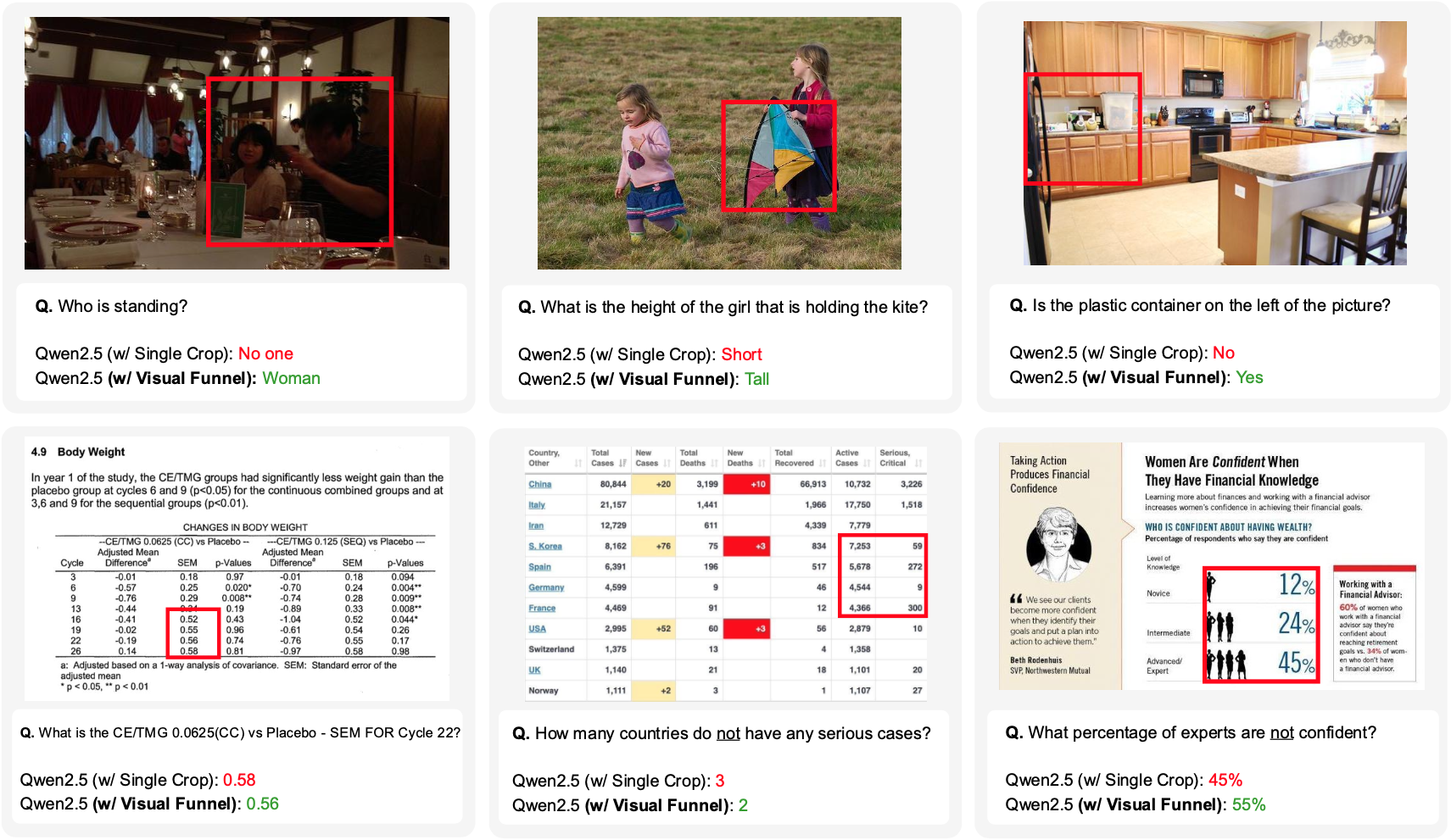}
    \caption{\textbf{Examples of Contextual Blindness.} Single-crop methods systematically remove essential context needed for correct reasoning, even when tight crops (red boxes) successfully isolate fine-grained details. \textbf{Top row:} (left) cropping only seated diners leads to overlooking the standing person; (center) excluding background reference objects results in incorrect height judgments; (right) different object positions within crop versus full image confuse spatial reasoning about left/right. \textbf{Bottom row:} (left) without surrounding column headers and labels, the model cannot identify which specific value corresponds to the requested metric; (center) excluding the ``Serious, Critical'' column header prevents identifying countries with empty cells in that column; (right) without seeing the question context about ``not confident,'' the model reports the visible ``45\%'' value instead of computing the inverse as 55\%. Even when provided with both the original image and the tight crop, MLLMs struggle to integrate information across these disparate scales, demonstrating the critical need for intermediate-scale representations that preserve hierarchical context. Results shown using Qwen2.5-VL-3B-Instruct.}
    \label{fig:contextual_blindness_examples}
\end{figure*}

\section{Introduction}
\label{sec:intro}

% PARAGRAPH1 : 문제 제기
Multimodal Large Language Models (MLLMs) have recently demonstrated significant capabilities in visual and language understanding. 
Despite this progress, a persistent challenge remains in perceiving small visual details, such as fine-grained text, distant object attributes, or subtle state differences. 
This \textit{small detail problem} serves as a major bottleneck, hindering their deployment in applications that require high precision.
%

% PARAGRAPH2 : 기존 방법론
Efforts to address this bottleneck generally follow a two-step paradigm: (1) \textbf{Localization}, identifying where the relevant detail is, and (2) \textbf{Integration}, determining how to structure and present that detail to the MLLM. 
Recent works have made significant strides in the Localization step by leveraging the inherent capabilities of MLLM to pinpoint areas of interest~\cite{wu2023vstar, zhang2025mllms}. 
These approaches either perform a multi-step, guided search to iteratively refine the region of interest, or directly analyze the model's internal signals, such as attention, in a single forward pass. 
Both philosophies have been effective in isolating the most salient region.

% PARAGRAPH3 : Contextual blindness
However, despite their differences in Localization strategies, these methods often rely on a simplistic approach for the crucial Integration step. 
Typically, a single, tightly-cropped high-resolution region is fed back into the model, sometimes alongside the original image. 
We identified that such naive integration introduces a critical limitation: while the model gains detail, it loses the intermediate context needed to interpret that detail. 
This overemphasis on an isolated region---despite the availability of global context---often fails due to \textbf{Contextual Blindness}, which we term as the underlying issue. 
As illustrated in Figure~\ref{fig:contextual_blindness_examples}, this issue is not due to missing pixels, but rather to a lack of \textit{structure}.
This observation leads to our central premise: ``what constrains MLLM's performance is not the absolute \textit{Quantity} of information, but the lack of \textit{Structural Diversity} in its input''. 
To address Contextual Blindness, we argue that a more sophisticated Integration strategy is required, one that goes beyond the limitations of single-crop methods.

% PARAGRAPH4 : 제안 방법
To this end, we introduce Visual Funnel (VF), a training-free methodology that holistically addresses both Localization and Integration in a holistic manner.
Drawing inspiration from the efficiency of internal signal-based methods, Visual Funnel enhances Localization while fundamentally rethinking the Integration process. It operates in two key steps:
\begin{enumerate}
    \item \textbf{Contextual Anchoring:} We first refine the Localization step by using a specialized ``search'' prompt to yield a more precise attention map from a single forward pass.
    \item \textbf{Entropy-Scaled Portfolio Generation:} For Integration, we leverage the attention map to construct a \textit{multi-scale information portfolio}. This portfolio dynamically adjusts crop sizes based on attention entropy and hierarchically refines crop centers, preserving the focal detail, its immediate surroundings, and the broader context.
\end{enumerate}

By explicitly addressing both Localization and Integration, Visual Funnel equips the MLLM with the structural diversity necessary to overcome Contextual Blindness. 
Our experiments validate this, demonstrating that Visual Funnel significantly outperforms baselines that rely on naive, single-crop integration. 
Additionally, we show that unstructured, repetitive information can be detrimental---introducing a `Redundancy Penalty'---further confirming that the hierarchical structure of our portfolio is the key to its success.

\section{Related Work}
\label{sec:related_work}
\subsection{Perception Limitations in Multimodal Large Language Models}
Recent Multimodal Large Language Models predominantly adopt modular pretrained architectures~\cite{li2023blip2,dai2023instructblip,liu2023llava,bai2023qwen}, connecting frozen Vision Transformers~\cite{dosovitskiy2021image} with Large Language Models via learnable connectors. This architecture inherently compresses visual information into a fixed number of tokens, creating a structural bottleneck for fine-grained perception, particularly for small visual elements.
Prior work has systematically demonstrated MLLMs' sensitivity to visual concept size~\cite{zhang2025mllms}, establishing a causal relationship between object size and perception accuracy. Related studies document additional perception failures including object hallucination~\cite{li2023hallucination} and visual blind spots~\cite{zhang2024exploring}. Recent surveys further analyze these challenges, highlighting weak spatial reasoning and poor fine-grained visual perception in MLLMs~\cite{jain2025words}. These limitations primarily address \textit{what} models fail to perceive, leaving \textit{how} visual information should be structured largely unexplored.
\subsection{High-Resolution Training for Enhanced Visual Perception}
Advances in high-resolution MLLM training have improved fine-grained visual understanding. LLaVA-NeXT~\cite{liu2024llavanext} introduces AnyRes for adaptive grid configurations. Qwen2-VL~\cite{wang2024qwen2vl} uses Naive Dynamic Resolution for variable visual tokens. InternVL2~\cite{chen2024internvl2} employs pixel shuffle with higher resolutions. Additional methods include patch-based strategies~\cite{li2024monkey}, cross-resolution fusion~\cite{wei2023vary}, and attention-efficient architectures~\cite{xu2024llavaud,tong2024cambrian}. Recent work scales vision pre-training to 4K resolution using selective processing of local regions~\cite{shi2025scaling}. While these approaches achieve performance gains, they require significant computational resources and apply static processing uniformly.
\subsection{Training-Free Inference-Time Enhancement}
Inference-time interventions leverage MLLMs' existing capabilities without modifications, categorized by external tool reliance. \\
\textbf{External Tool-Based Approaches} use specialized vision models for visual attention. V*~\cite{wu2023vstar} refines regions iteratively with tools like YOLO~\cite{redmon2016you} and SAM~\cite{kirillov2023segment}. Visual programming~\cite{gupta2023visual,suris2023vipergpt} orchestrates tools via code. Set-of-Mark~\cite{yang2023set} overlays spatial marks on segmented regions. These require multiple passes and external dependencies. \\
\textbf{Internal Signal-Based Approaches} exploit internal representations without tools. Compositional chain-of-thought~\cite{mitra2024ccot} uses scene graphs for compositional knowledge. ControlMLLM~\cite{wu2024controlmllm} optimizes latent variables to control attention for region description. ViCrop~\cite{zhang2025mllms} analyzes attention patterns to identify salient regions based on MLLMs' sensitivity to visual subject size~\cite{zhang2024perceiving}. However, by providing only a single tight crop alongside the original image, it loses intermediate-scale context essential for relational reasoning---a limitation we term \textit{contextual blindness}.
We address this by constructing adaptive multi-scale visual portfolios that preserve contextual hierarchies, as detailed in Section~\ref{sec:method}.
\begin{figure*}[!t]
  \centering
     \includegraphics[width=1.0\linewidth]{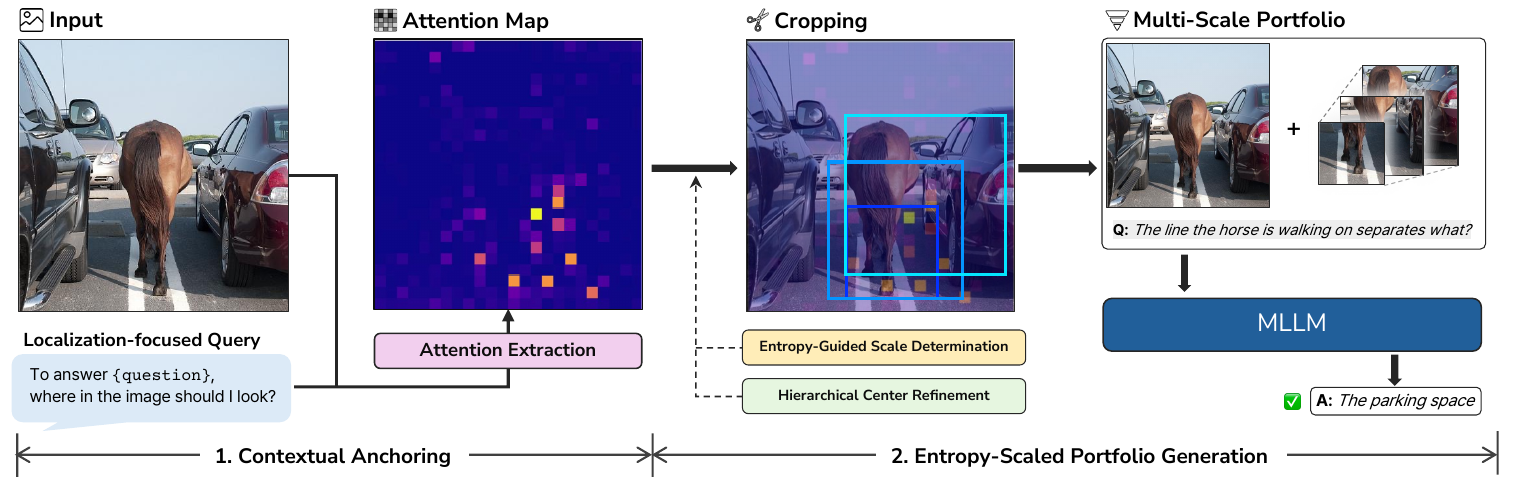}
    \caption{
    \textbf{An overview of our proposed Visual Funnel methodology}. Visual Funnel resolves Contextual Blindness through a two-step, training-free process.
    \textbf{(1) Contextual Anchoring:} A localization-focused query guides the MLLM to establish a semantic anchor by generating a precise spatial attention map for the region of interest.
    \textbf{(2) Entropy-Scaled Portfolio Generation:} This attention map then serves as the foundation for generating a multi-scale information portfolio. Crop sizes are dynamically scaled based on attention entropy, and their positions are hierarchically refined to preserve crucial context.
    This final portfolio, comprising the original image and multiple contextual crops, provides the MLLM with the necessary structural diversity to correctly answer the question (e.g., identifying the context as a `parking space` instead of a `lane`), a task where naive single-crop methods often fail.
    }
    \vspace{-3mm}
    \label{fig:full_pipeline}
\end{figure*}

\section{Method}
\label{sec:method}
To address the challenge of perceiving small details in MLLMs, previous works have advanced localization through attention-guided cropping~\cite{zhang2025mllms} or iterative search~\cite{wu2023vstar}. 
However, these methods rely on a single-scale crop for perception, discarding intermediate context and creating a structural disconnect between global view and focal detail. 
We formalize this critical limitation as \textbf{Contextual Blindness} where coexisting global context and focal detail remain unbridgeable due to the lack of intermediate scales.

We propose Visual Funnel, which is illustrated in Figure~\ref{fig:full_pipeline} as a solution to Contextual Blindness through adaptive multi-scale information portfolios. 
We begin by analyzing the problem (Section~\ref{sec:contextual_blindness}), then describe context-aware attention extraction (Section~\ref{sec:attention_localization}) and Entropy-Scaled Portfolio Generation through hierarchical center refinement and entropy-guided scaling (Section~\ref{sec:multiscale}).

\subsection{Contextual Blindness}
\label{sec:contextual_blindness}

MLLMs often struggle to perceive small visual details in images. 
Previous approaches address this through a two-step paradigm: (1) \textit{Localization}, identifying where to focus, and (2) \textit{Integration}, enabling the model to recognize details in that region. 
Recent attention-based methods~\cite{zhang2025mllms} have demonstrated strong localization capability, while iterative search approaches~\cite{wu2023vstar} achieve precise localization through multiple forward passes. 
These methods suggest that the primary bottleneck is not localization, but integration.

Current integration strategies, such as providing the best single crop alongside the original image~\cite{zhang2025mllms}, rely on the core assumption that coexisting global context (the original image) and focal detail (the crop) are sufficient for accurate perception. 
However, we often observe failure cases where contextual understanding is required.
Even with both a global view and focal detail, the MLLM cannot answer correctly. 
The root cause of this failure is the missing intermediate-scale context that bridges the two.
We identify and term this underlying issue as ``Contextual Blindness'': a failure of reasoning that occurs even when the MLLM possesses all necessary visual information (i.e., in the global image and the focal crop), simply because the structural disconnect between the focal detail and the global context makes them unbridgeable. 
This ``blindness'' results in critical failures for robust visual understanding, as illustrated in Figure~\ref{fig:contextual_blindness_examples}.

For instance, an object's attribute may be relative rather than absolute, defined by its relation to the surrounding context.
In Figure~\ref{fig:contextual_blindness_examples}, answering the question regarding the kite-holding girl's height (\textit{Short}) requires comparing her to the other girl in the frame. 
A single-scale crop focused only on the target girl would sever this relational link, causing the model to lose the necessary reference point and potentially misinterpret her height.
Similarly, such failure occur when an answer requires synthesizing information from spatially distinct pieces of chart or table.
To correctly answer the question regarding the percentage of experts who are \textit{not confident} (Figure~\ref{fig:contextual_blindness_examples}), the MLLM must first locate the `Advanced/Expert` category, associate it with the corresponding `45\%` value from the bar chart, and understand that this value represents those who \textit{are} confident, before correctly processing the negation in the query. 
A single crop focused on either the text or the bar chart in isolation would fragment this information, making the compositional reasoning required for this task impossible.
In such cases, the MLLM does not lack information---the necessary pixels are available---but rather lacks the correct information \textit{structure} to process it correctly.

\subsection{Visual Funnel}
\label{sec:vf}

To overcome Contextual Blindness, We propose Visual Funnel, a training-free approach that constructs adaptive multi-scale information portfolios. 
Visual Funnel dynamically determines both crop locations and the extent of context required, using hierarchical refinement and entropy-guided scaling, respectively.
It provides complementary views across focal, immediate, and broader context scales, bridging fine-grained details with the global structure.

\subsubsection{Step 1: Contextual Anchoring}
\label{sec:attention_localization}

To identify focal regions requiring detailed examination, we extract attention maps from the MLLM's internal representations.
Unlike direct answering, which may produce hallucinations when visual information is insufficient, we prompt the MLLM with a localization-focused query: \textit{'To answer `\{question\}', where in the image should I look?'}.
This query encourages the model to identify the \textit{region} containing relevant information, rather than prematurely committing to an answer when details are unclear.
\\
\\
\noindent\textbf{Attention Extraction.}
Following the approach in ViCrop~\cite{zhang2025mllms}, we extract cross-attention from the same pre-specified layer used in ViCrop for each backbone during a single forward pass.
Given an image-question pair $(I, q)$, modern MLLMs process the image through a Vision Transformer (ViT) encoder into $N \times N$ patch tokens, which are then projected into $T$ image tokens for the LLM. 
We then extract the softmax cross-attention of the first response token to all image tokens, yielding $\mathbf{A}(I, q) \in \mathbb{R}^{H \times 1 \times T}$ where $H$ is the number of attention heads. 
Averaging over the heads gives the attention map:
\begin{equation}
\hat{\mathbf{A}}(I, q) = \frac{1}{H} \sum_{h=1}^{H} \mathbf{A}^h(I, q) \in \mathbb{R}^{1 \times T}.
\end{equation}

For MLLMs using Transformer-based connectors (\eg, InstructBLIP), we combine the LLM-to-token attention with the connector's token-to-patch cross-attention to establish spatial correspondence. 
For MLLMs with direct projection (\eg, LLaVA-1.5), image tokens directly correspond to spatial patches, producing a spatial attention map $\mathbf{A} \in \mathbb{R}^{B_h \times B_w}$ over ViT output patches. 
We normalize $\mathbf{A}$ to form a probability distribution:
\begin{equation}
\mathbf{A}_{\text{norm}}[i,j] = \frac{\mathbf{A}[i,j]}{\sum_{i',j'} \mathbf{A}[i',j']},
\end{equation}
where $\mathbf{A}_{\text{norm}} \in \mathbb{R}^{B_h \times B_w}$ represents the probability that spatial block $(i,j)$ contains information relevant to answering the question. 
This attention map serves as the foundation for our Entropy-Scaled Portfolio Generation (Section~\ref{sec:multiscale}), requiring only a single forward pass without architectural modifications.

\subsubsection{Step 2: Entropy-Scaled Portfolio Generation}
\label{sec:multiscale}

Given the spatial attention map $\mathbf{A}_{\text{norm}}$ from Section~\ref{sec:attention_localization}, we construct an adaptive multi-scale portfolio through two interleaved mechanisms: \textit{entropy-guided scaling} determines \textit{how much} context each crop requires based on attention uncertainty, while \textit{hierarchical center refinement} determines \textit{where} to center crops to handle asymmetric attention distributions.

\noindent\textbf{Entropy-Guided Scale Determination.}
We observe that attention entropy directly correlates with the contextual requirements of a region. 
Low entropy ($H \approx 0$) indicates highly confident, localized attention, requiring minimal additional context.
High entropy ($H \approx \log |\mathbf{A}|$) suggests diffuse attention, indicating ambiguity or relationships between multiple elements that demand broader context to resolve uncertainty.

We compute the normalized Shannon entropy over the spatial attention distribution:
\begin{equation}
H_{\text{norm}}(I, q) = -\frac{1}{\log(B_h \cdot B_w)} \sum_{i,j} \mathbf{A}_{\text{norm}}[i,j] \log \mathbf{A}_{\text{norm}}[i,j],
\end{equation}
where $H_{\text{norm}} \in [0, 1]$ quantifies attention uncertainty. 
We define adaptive expansion factors as linear functions of $H_{\text{norm}}$:
\begin{align}
\alpha_1(I, q) &= 1.2 + 0.6 \cdot H_{\text{norm}}(I, q) \in [1.2, 1.8], \\
\alpha_2(I, q) &= 1.6 + 1.2 \cdot H_{\text{norm}}(I, q) \in [1.6, 2.8],
\end{align}
These hyperparameters were empirically determined. 
Our analysis shows that the model's performance remains stable across a reasonable range of these values, indicating that our method is not sensitive to specific choices. 

These mappings ensure minimal context expansion (1.2$\times$, 1.6$\times$) even for confident attention---preventing Contextual Blindness by always providing intermediate context---while allowing aggressive expansion (1.8$\times$, 2.8$\times$) for uncertain cases requiring broader relationships. 

\noindent\textbf{Hierarchical Center Refinement.}
In standard multi-scale approaches, attention is assumed to be centered within crops, but attention distributions are often asymmetric. 
In document images, target cells may lie near table edges, and in outdoor scenes, salient objects may occupy corners.
We address this issue through hierarchical refinement where each level's center is computed based on attention within its parent crop region.

Starting from the global attention center $\boldsymbol{\mu}_0$ over the entire image, we iteratively refine the center at each scale level $\ell$ within the region $\mathcal{R}_\ell$ defined by the previous level's crop:
\begin{equation}
\boldsymbol{\mu}_\ell(I, q) = \frac{\sum_{(i,j) \in \mathcal{R}_\ell} \mathbf{c}_{ij} \cdot \mathbf{A}_{\text{norm}}[i,j]}{\sum_{(i,j) \in \mathcal{R}_\ell} \mathbf{A}_{\text{norm}}[i,j]}
\end{equation}
where $\mathbf{c}_{ij} \in \mathbb{R}^2$ denotes the center coordinate of spatial block $(i,j)$ in image space. 
This refinement automatically corrects for asymmetry: if attention within a crop is skewed toward one edge, $\boldsymbol{\mu}_{\ell}$ shifts in that direction relative to $\boldsymbol{\mu}_{\ell-1}$, ensuring the next scale is optimally positioned to capture relevant context.

\noindent\textbf{Multi-Scale Portfolio.}
Let $S$ denote the MLLM's input image resolution. 
Our final portfolio consists of three crops, each centered at its hierarchically refined location and scaled according to entropy:
\begin{itemize}[leftmargin=*,itemsep=0pt,topsep=3pt]
\item \textbf{Crop}\textsubscript{\textbf{focal}}: $S \times S$ pixels centered at $\boldsymbol{\mu}_0$ \textit{(focal detail)}
\item \textbf{Crop}\textsubscript{$\boldsymbol{\alpha_1}$}: $(\alpha_1 \cdot S) \times (\alpha_1 \cdot S)$ pixels centered at $\boldsymbol{\mu}_1$ \textit{(immediate context)}
\item \textbf{Crop}\textsubscript{$\boldsymbol{\alpha_2}$}: $(\alpha_2 \cdot S) \times (\alpha_2 \cdot S)$ pixels centered at $\boldsymbol{\mu}_2$ \textit{(broader context)}
\end{itemize}
Each crop is resized to $S \times S$ pixels, encoded by the MLLM's vision encoder, and concatenated with the original image tokens. The MLLM then processes this enriched token sequence—comprising global context (original image) and multi-scale focal information (three crops)—to generate the final answer.
% \definecolor{mediumgray}{gray}{0.8}
%Note to anyone using this template in the future ? will give you bold vertical lines, while | is regular
% % Adjust row height
\renewcommand{\arraystretch}{1.1}
\newcolumntype{?}{!{\vrule width 2 pt}}
\newcolumntype{P}[1]{>{\centering\arraybackslash}p{#1}}
\newcolumntype{G}[1]{>{\columncolor{lightgray}\centering\arraybackslash}p{#1}}
% \begin{center}
\begin{table*}[t]
\begin{center}
\begin{tabular}{m{0.22\textwidth}G{0.08\textwidth}G{0.08\textwidth}G{0.09\textwidth}G{0.08\textwidth}P{0.08\textwidth}P{0.08\textwidth}P{0.08\textwidth}}
        \multicolumn{1}{c}{}& \multicolumn{4}{c}{\textbf{Grounded Visual QA}} & \multicolumn{3}{c}{\textbf{Recognition Visual QA}} \\
        \toprule
        Model & TextVQA & GQA & DocVQA & InfoVQA & POPE & AOKVQA & VQAv2  \\ \hline
        LLaVA-1.5-7B & 47.9 & 60.1 & 15.9 & 12.0 & 85.6 & 58.7 & 75.4  \\
        \qquad w/ViCrop & 54.1 & 60.4 & 19.4 & 12.6 & 87.4 & 60.4 & 76.1 \\
        \qquad w/ViCrop (Top-3) & 53.5 & 60.5 & 19.2 & 12.9 & 87.5 & 60.6 & 76.6 \\
        \qquad \textbf{w/Visual Funnel} & \bf{59.1}\gcol{+11.2} & \bf{61.3}\gcol{+1.2}& \bf{22.8}\gcol{+7.0}&\bf{15.1}\gcol{+3.1} & \bf{88.3}\gcol{+2.7}&\bf{60.6}\gcol{+1.9} & \bf{76.7}\gcol{+1.3}  \\ \hline
        InstructBLIP-7B & 33.4 & 49.4 & 9.2 & 12.8 & 84.7 & 59.9 & 76.3 \\ 
        \qquad w/ViCrop & 45.3 & 49.7 & 9.9 & 15.8 & 86.6 & 61.3 & 76.8 \\
        \qquad w/ViCrop (Top-3) & 45.8 & 49.8 & 10.1 & 16.0 & 87.0 & 61.5 & 77.1 \\
        \qquad \textbf{w/Visual Funnel} & \bf{49.8}\gcol{+16.4} &\bf{50.6}\gcol{+1.2}&\bf{18.5}\gcol{+9.3}&\bf{25.1}\gcol{+12.3} & \bf{87.1}\gcol{+0.5} & \bf{61.6}\gcol{+1.7} & \bf{77.2}\gcol{+0.9}\\ \hline
        Qwen2.5-VL-3B-Instruct & 70.1 & 61.2 & 51.5 & 34.2 & 87.1 & 57.9 & 78.9 \\ 
        \qquad w/ViCrop & 76.0 & 60.8 & 54.2 & 39.4 & 88.4 & 59.4 & 78.2 \\
        \qquad w/ViCrop (Top-3) & 76.7 & 61.4 & 55.3 & 39.9 & 88.5 & 60.3 & 79.4 \\
        \qquad \textbf{w/Visual Funnel} & \bf{79.8}\gcol{+9.7}  & \bf{62.2}\gcol{+1.0} & \bf{61.1}\gcol{+9.6}& \bf{49.6}\gcol{+15.4} & \bf{88.5}\gcol{+1.4} & \bf{60.4}\gcol{+2.5}& \bf{79.5}\gcol{+0.6}
        \\\bottomrule
    \end{tabular}
    \end{center}
\vspace{-3mm}
\caption{
\textbf{Main results of our proposed Visual Funnel method.}
    The benchmarks are categorized into \textit{Grounded Visual QA}, which is highly sensitive to Contextual Blindness, and standard \textit{Recognition Visual QA}.
    We report zero-shot accuracy (\%). Numbers in parentheses denote the absolute performance gain of Visual Funnel over the \texttt{Base MLLM (No Cropping)} baseline.
    The best results for each model are shown in \textbf{bold}. Our method consistently and significantly outperforms all baselines, with the most substantial gains observed on the detail-oriented Grounded Visual QA tasks.
}
\label{tab:main_results}
\end{table*}
% \end{center}

\section{Experiments}
\label{sec:exp}

% \subsection{Implementation Details}
% \label{sec:implementation_details}

We implement our proposed Visual Funnel methodology in PyTorch. As Visual Funnel is a training-free, inference-time approach, our implementation only requires the standard infrastructure necessary for running inference on the base MLLMs. For all experiments, we use the official, publicly available implementations of the models evaluated, as detailed in Section~\ref{sec:models}.

% For all applications of Visual Funnel, the final enriched input provided to the MLLM consists of the original image tokens concatenated with the tokens from the three generated crops in our multi-scale portfolio. Further details on prompt formulations and the hyperparameter settings used for entropy-guided scaling are provided in Appendix.

\subsection{Datasets}
We evaluate our method on seven widely-used VQA benchmarks, categorized into two groups based on evaluation focus.

\minisection{Grounded Visual QA} This category emphasizes fine-grained visual perception requiring models to recognize and reason about small textual or visual details. We evaluate on: (i)~\textbf{TextVQA}~\cite{singh2019textvqa}, containing 45K questions requiring scene text reading without external OCR tokens; (ii)~\textbf{GQA}~\cite{hudson2019gqa}, with 22M compositional questions on Visual Genome scene graphs; (iii)~\textbf{DocVQA}~\cite{mathew2021docvqa}, featuring 50K questions on document understanding with complex layouts; (iv)~\textbf{InfoVQA}~\cite{mathew2022infographicvqa}, comprising 5.4K questions on infographics with charts and diagrams.

\minisection{Recognition Visual QA} This category assesses broader visual recognition, spatial reasoning, and knowledge-based understanding. We evaluate on: (i)~\textbf{POPE}~\cite{li2023hallucination}, evaluating object hallucination through binary presence questions; (ii)~\textbf{A-OKVQA}~\cite{schwenk2022aokvqa}, requiring commonsense and world knowledge integration; (iii)~\textbf{VQAv2}~\cite{goyal2017vqav2}, containing 1M+ questions with balanced answer distributions.

\subsection{Models}
\label{sec:models}
We apply Visual Funnel to three representative MLLMs with different architectural designs, demonstrating generalizability across varying model capacities.

% \minisection{LLaVA}
% The LLaVA series~\cite{liu2023llava} pioneered a simple yet effective approach to vision-language instruction tuning, employing a linear projection to map CLIP visual features into the LLM's token space. This architecture enables end-to-end training on diverse instruction-following datasets while maintaining computational efficiency. LLaVA-1.5~\cite{liu2023llava15} enhances this design by replacing the linear projection with a two-layer MLP and incorporating broader pretraining datasets. In our experiments, we evaluate LLaVA-1.5 with Vicuna-7B backbone, which uses CLIP ViT-L/14 as visual encoder at 336$\times$336 resolution, producing 576 visual tokens.

\minisection{LLaVA}
The LLaVA series~\cite{liu2023llava} pioneered a simple yet effective approach to vision-language instruction tuning, employing a linear projection to map CLIP visual features into the LLM's token space. LLaVA-1.5 enhances this design by replacing the linear projection with a two-layer MLP. In our experiments, we evaluate \texttt{LLaVA-1.5-7B}, which uses a CLIP ViT-L/14 visual encoder at a $336 \times 336$ input resolution, producing 576 visual tokens.

% \minisection{InstructBLIP}
% Built on BLIP-2~\cite{li2023blip2}, InstructBLIP~\cite{dai2023instructblip} employs a Q-Former architecture that compresses visual information through learnable query tokens while maintaining instruction awareness. The Q-Former cross-attends to both visual features from the frozen encoder and task prompts simultaneously, creating context-dependent visual representations. We evaluate the Vicuna-7B variant, which uses 32 learnable queries to produce instruction-aware visual tokens, demonstrating strong performance on instruction-following benchmarks.

\minisection{InstructBLIP}
Built on BLIP-2~\cite{li2023blip2}, InstructBLIP~\cite{dai2023instructblip} employs a Q-Former architecture that compresses visual information through learnable query tokens. The Q-Former cross-attends to both visual features and task prompts simultaneously, creating context-dependent visual representations. We evaluate the Vicuna-7B variant, which uses 32 learnable queries and a default input resolution of $224 \times 224$.

% \minisection{Qwen-VL}
% The Qwen-VL family~\cite{bai2023qwen} introduces versatile vision-language models capable of understanding, localization, and text reading. Qwen2-VL~\cite{wang2024qwen2vl} advances this with Naive Dynamic Resolution, which adaptively processes images at their native aspect ratios by converting them into variable numbers of visual tokens, supporting resolutions up to 4K during inference. We evaluate Qwen2.5-VL-3B-Instruct, a compact instruct-tuned variant that maintains strong capabilities in document understanding and fine-grained recognition despite its smaller parameter count, making it suitable for efficient deployment.

\minisection{Qwen-VL}
The Qwen-VL family~\cite{bai2023qwen} introduces versatile vision-language models capable of understanding, localization, and text reading. Qwen2-VL~\cite{wang2024qwen2vl} advances this with Naive Dynamic Resolution, which adaptively processes images at their native aspect ratios. We evaluate \texttt{Qwen2.5-VL-3B-Instruct}, a compact instruct-tuned variant suitable for efficient deployment.

\subsection{Baselines}
\label{sec:baselines}
To evaluate the effectiveness of Visual Funnel, we compare its performance against three key baselines. These are presented in our main results in Table~1 and are designed to isolate the impact of our structured, multi-scale approach.

\minisection{Base MLLM (No Cropping)} Our first baseline is the standard zero-shot performance of each base MLLM. The model is applied directly to the benchmark questions without any cropping or other interventions. This baseline establishes the lower-bound performance and serves as the reference point for quantifying the improvements gained from any cropping-based strategy.

\minisection{w/ViCrop} Our primary baseline for comparison is the single-crop enhancement method inspired by ViCrop~\cite{zhang2025mllms}, which we denote as \texttt{w/ViCrop}. This approach leverages the model's internal attention map from a single forward pass to identify the most salient visual region relevant to the query. A single tight crop of this region is then resized and fed back to the model alongside the original image. This represents the standard, state-of-the-art for single-crop, training-free enhancement methods.

\minisection{w/ViCrop (Top-3)} To specifically test our core hypothesis that `Structural Diversity $>$ Quantity', we designed a crucial multi-crop baseline, termed \texttt{w/ViCrop (Top-3)}. This baseline uses the same attention map generated for \texttt{w/ViCrop}. However, instead of selecting only the single most salient region, it identifies the top three non-overlapping regions with the highest average attention scores. These three crops are then individually resized and concatenated with the original image tokens. Crucially, this baseline provides the same number of additional crops as our Visual Funnel, allowing us to isolate the effect of our portfolio's \textit{hierarchical structure} from the simple effect of adding more visual tokens. It directly challenges our method by providing an equal `Quantity' of unstructured information.

Finally, we compare these baselines against our proposed \texttt{w/Visual Funnel} method, which also uses three crops but, unlike \texttt{w/ViCrop (Top-3)}, constructs them into an adaptive, hierarchical portfolio as detailed in Section~\ref{sec:method}.

\subsection{Results}

We evaluate Visual Funnel on three representative MLLMs---LLaVA-1.5-7B, InstructBLIP-7B, and Qwen2.5-VL-3B---across seven benchmarks, categorized into Grounded Visual QA and Recognition Visual QA. The full results are presented in Table~\ref{tab:main_results}.

\subsubsection{Analysis on Grounded Visual QA}

Grounded Visual QA (TextVQA, DocVQA, InfoVQA, GQA) is the primary testbed for \textit{Contextual Blindness}, which requires reasoning about the relationship between fine-grained details and their surrounding context. As shown in Table~\ref{tab:main_results}, Visual Funnel demonstrates significant performance gains over all baselines on these datasets, with improvements ranging from +9.3 to +16.4 points on DocVQA and InfoVQA across different models.

Crucially, this analysis allows us to validate our core hypothesis: \textit{Structural Diversity $>$ Quantity}. We designed the \texttt{w/ViCrop (Top-3)} baseline to isolate the impact of `Quantity' (adding more detail tokens) from `Structural Diversity' (adding multi-scale hierarchical context). Both \texttt{w/ViCrop (Top-3)} and \texttt{w/Visual Funnel} use identical token budgets and extract crops from the same attention map, differing only in the structural organization of information.

The results for \texttt{w/ViCrop (Top-3)} are striking. For Qwen2.5-VL, adding more crops provides only negligible improvements over the standard \texttt{w/ViCrop} (\textit{e.g.}, TextVQA: 76.0 $\rightarrow$ 76.7, +0.7). More notably, for LLaVA-1.5, \texttt{w/ViCrop (Top-3)} consistently performs \textit{worse} than the single-crop \texttt{w/ViCrop} (TextVQA: 54.1 $\rightarrow$ 53.5, $-$0.6; DocVQA: 19.4 $\rightarrow$ 19.2, $-$0.2). This ``Redundancy Penalty'' strongly indicates that adding unstructured, repetitive information can be actively detrimental to the MLLM's reasoning.
In sharp contrast, our Visual Funnel consistently and significantly outperforms both baselines (\textit{e.g.}, LLaVA/TextVQA: 59.1 vs. 53.5, +5.6; InstructBLIP/DocVQA: 18.5 vs. 10.1, +8.4). This confirms that the hierarchical multi-scale structure (focal $\rightarrow$ immediate $\rightarrow$ broader) is the key factor in resolving these complex reasoning tasks.

We note that GQA shows more modest gains (+1.0 to +1.2). This is consistent with our hypothesis, as GQA questions often feature `Larger Visual Concepts' in natural scenes, which rely less on the intermediate-scale context that defines \textit{Contextual Blindness}.

\subsubsection{Analysis on Recognition Visual QA}

On standard recognition benchmarks (POPE, AOKVQA, VQAv2), where \textit{Contextual Blindness} is not the primary bottleneck, \texttt{w/Visual Funnel} shows modest improvements over \texttt{w/ViCrop} (averaging +0.5 to +1.0 points). These gains are substantially smaller than those observed on Grounded Visual QA (averaging +7.1 to +12.7 points).

This finding reinforces our core claim. In the absence of complex requirements for perceiving small visual details with surrounding context, the unique advantage of VF's `Structural Diversity' is less pronounced. This demonstrates that VF's significant advantage is precisely targeted at solving the small-concept \textit{Contextual Blindness} it was designed to address, while maintaining robust performance elsewhere. The differential effectiveness across benchmark categories directly validates our problem formulation: Visual Funnel is not a generic performance enhancer but a targeted solution for \textit{Contextual Blindness} in small detail perception.

\subsection{Ablations}

We conduct a comprehensive ablation study using the \texttt{Qwen2.5-VL-3B-Instruct} model to systematically dissect the contributions of the core components within Visual Funnel. Our analysis is designed to isolate the distinct effects of our two main steps: (1) \textit{Step 1: Contextual Anchoring}, and (2) \textit{Step 2: Entropy-Scaled Portfolio Generation}. The results are summarized in Table~\ref{tab:ablation_components}.

\minisection{The Limitation of Contextual Anchoring Alone} When we ablate our portfolio construction (\texttt{Visual Funnel w/o Step 2}), applying only our specialized localization prompt (Step 1) to a single-crop baseline, we observe only a marginal performance increase (+0.9 on DocVQA) over the standard \texttt{ViCrop} baseline. This finding is critical: it demonstrates that even a more precise attention map is insufficient to overcome the core issue. This strongly supports our claim that Contextual Blindness is a fundamental problem of information \textit{structure}, which is addressed by our Step 2, not merely localization accuracy from Step 1.

\minisection{The Decisive Impact of Entropy-Scaled Portfolio Generation} Conversely, when we ablate our specialized prompt (\texttt{Visual Funnel w/o Step 1}) and apply our portfolio construction method (Step 2) to a standard, less-precise attention map, the model dramatically outperforms the \texttt{ViCrop} baseline (+5.6 on DocVQA). This result provides direct evidence for our central hypothesis that providing `Structural Diversity' is the key to resolving complex visual reasoning tasks. The performance gains are primarily driven by the adaptive and hierarchical structure of the portfolio generated in Step 2.

\minisection{Synergistic Effect of Both Steps} Finally, our full \texttt{Visual Funnel (Ours)} model, which integrates both steps, achieves the best results. It consistently outperforms all other variants, indicating a clear synergistic effect. The high-quality attention map from Step 1 provides a more robust anchor for the portfolio construction in Step 2, and the portfolio in turn provides the necessary structural context that a single crop cannot. This comprehensive analysis validates our two-step design, confirming that both components are essential and work in concert to effectively resolve Contextual Blindness.

Additionally, we conducted further analyses on the optimal number of portfolio crops and a detailed efficiency comparison (e.g., input tokens, latency) against the \texttt{ViCrop (Top-3)} baseline.

% \begin{table}[t]
% \centering
% \resizebox{\columnwidth}{!}{%
% \begin{tabular}{l|ccc}
% \hline
% \textbf{Configuration}              & \textbf{DocVQA} & \textbf{InfoVQA} \\ \hline
% ViCrop (Baseline)                   & 54.2          & 39.4           \\
% Visual Funnel w/o Step 2            & -          & -           \\
% Visual Funnel w/o Step 1            & -          & -           \\
% \textbf{Visual Funnel (Ours)}       & 61.1          & 49.6           \\ \hline
% \end{tabular}%
% }
% \caption{Ablation study examining the robustness of QEVA to different combinations of Video-LMM and LLM models compared to the default setting (Gemini-1.5 Pro + GPT-4o). We report correlations with human annotations (Kendall's $\tau_b$, $\tau_c$, and Spearman's $\rho$) when replacing original models with alternative open-source models (Qwen2.5-VL, InternVL3, LLaMA-3.1, Gemma-3).
% \textbf{Takeaway:} QEVA maintains high correlations with human judgments even when using alternative open-source models, indicating practical applicability and cost-effectiveness without relying solely on costly API-based models.}
% \label{tbl:component-ablation}
% \end{table}

\begin{table}[t]
    \centering
    \begin{tabular}{l|cc}
        \toprule
        \textbf{Configuration} & \textbf{DocVQA} & \textbf{InfoVQA} \\
        \midrule
        ViCrop (Baseline) & 54.2 & 39.4 \\
        Visual Funnel w/o Step 2 & 55.1 & 40.3 \\ % TODO: Replace with your actual results
        Visual Funnel w/o Step 1 & 59.8 & 47.9 \\ % TODO: Replace with your actual results
        \bfseries Visual Funnel (Ours) & \bfseries 61.1 & \bfseries 49.6 \\
        \bottomrule
    \end{tabular}
    \caption{Ablation study on the core components of Visual Funnel, conducted with the \texttt{Qwen2.5-VL-3B-Instruct} model. We report accuracy on DocVQA and InfoVQA.}
    \label{tab:ablation_components}
\end{table}
\section{Conclusion}
\label{sec:conclusion}

In this work, we identified and addressed ``Contextual Blindness,'' a critical failure mode in Multimodal Large Language Models where the structural disconnect between global context and focal detail impedes fine-grained visual reasoning. We argued that resolving this issue requires moving beyond naive, single-crop integration methods. Our approach is rooted in the central premise that for complex perception, `Structural Diversity' is more critical than the mere `Quantity' of visual information. To this end, we proposed Visual Funnel, a training-free, two-step methodology that operationalizes this principle. By first employing a specialized prompt for robust attention-guided localization and then, crucially, constructing an adaptive multi-scale information portfolio, Visual Funnel provides MLLMs with the necessary hierarchical context to bridge the gap between detail and context. Through extensive experiments on several MLLMs and challenging benchmarks, we demonstrated that Visual Funnel significantly outperforms existing methods. Our ablation studies further validated our design, revealing a `Redundancy Penalty' for unstructured multi-crop inputs and confirming that the hierarchical structure of our portfolio is the key to its success.

\section{Limitations}
\label{sec:limitations}

While Visual Funnel demonstrates strong performance, we acknowledge several limitations.

First, the effectiveness of our method is predicated on a reasonably accurate initial attention map from Step 1. In rare cases where the MLLM completely fails to localize the region of interest, the quality of the generated portfolio may be compromised.

Second, our current approach is designed to resolve questions centered around a single region of interest. Consequently, it may not be suitable for complex queries that require synthesizing information from multiple, spatially distinct focal points simultaneously.

Finally, while our method is training-free, it introduces a computational overhead at inference time due to the processing of multiple image crops. Although our experiments suggest this is a favorable trade-off for the substantial accuracy gains on detail-oriented tasks, this overhead could be a concern in latency-sensitive scenarios.

\section*{Acknowledgments}
This work was partly supported by Institute of Information and Communications Technology Planning and Evaluation (IITP) grant funded by the Korea Government (MSIT) (No. RS-2022-II220184, Development and Study of AI Technologies to Inexpensively Conform to Evolving Policy on Ethics) and partly supported by the Institute of Information and Communications Technology Planning and Evaluation (IITP) grant funded by the Korea Government (MSIT) [RS-2021-II211341, Artificial Intelligence Graduate School Program (Chung-Ang University)].

{
    \small
    \bibliographystyle{ieeenat_fullname}
    \bibliography{main}
}

\appendix
\maketitlesupplementary
\appendix

\section{Hyperparameter Sensitivity Analysis}
\label{sec:appendix_hyperparams}

In Section 3.2.2 of the main paper, we introduced the \textit{Entropy-Guided Scale Determination} mechanism. The crop expansion factors, $\alpha_1$ (for immediate context) and $\alpha_2$ (for broader context), are computed as linear functions of the normalized attention entropy $H_{\text{norm}}$:

\begin{equation}
    \alpha_k(I, q) = \beta_k + \gamma_k \cdot H_{\text{norm}}(I, q)
    \label{eq:generalized_alpha}
\end{equation}

where $\beta_k$ represents the base expansion factor (minimum context size) and $\gamma_k$ represents the sensitivity coefficient to the model's uncertainty. Our default configuration uses $\mathcal{C}_{\text{default}} = \{ \beta_1=1.2, \gamma_1=0.6, \beta_2=1.6, \gamma_2=1.2 \}$.

To demonstrate that our method is robust and not overfitted to specific ``magic numbers,'' we conduct a comprehensive sensitivity analysis using the Qwen2.5-VL-3B-Instruct backbone on the DocVQA dataset. Importantly, these default parameters were determined using a small held-out validation set from GQA and kept fixed across all benchmarks reported in the main paper.

\subsection{Impact of Entropy Sensitivity ($\gamma$)}
First, we investigate the necessity of the adaptive scaling mechanism. We vary the sensitivity coefficients $\gamma_1$ and $\gamma_2$ while keeping the base factors $\beta$ fixed. Setting $\gamma=0$ represents a \textit{Static} baseline where crop sizes are fixed regardless of attention uncertainty.

As shown in Table~\ref{tab:sensitivity_gamma}, the adaptive configuration ($\gamma > 0$) consistently outperforms the static approach. The performance peaks at our default setting but remains stable within a reasonable range ($\gamma_1 \in [0.4, 0.8]$), confirming that allocating broader context to uncertain regions is crucial for resolving Contextual Blindness.

\begin{table}[h]
    \centering
    \small
    \setlength{\tabcolsep}{8pt}
    \resizebox{\linewidth}{!}{
    \begin{tabular}{lcccc}
        \toprule
        Configuration & $\gamma_1$ & $\gamma_2$ & DocVQA Acc. (\%) & $\Delta$ \\
        \midrule
        Static (Fixed Size) & 0.0 & 0.0 & 59.5 & -1.6 \\
        Weak Adaptation & 0.3 & 0.6 & 60.4 & -0.7 \\
        \textbf{Default (Ours)} & \textbf{0.6} & \textbf{1.2} & \textbf{61.1} & \textbf{--} \\
        Strong Adaptation & 0.9 & 1.8 & 60.8 & -0.3 \\
        \bottomrule
    \end{tabular}}
    \caption{\textbf{Ablation on Entropy Sensitivity.} We analyze the impact of the sensitivity coefficient $\gamma$. The results validate that adaptive scaling based on attention entropy yields better performance than static cropping ($\gamma=0$).}
    \label{tab:sensitivity_gamma}
\end{table}

\subsection{Robustness of Base Expansion Factors ($\beta$)}
Next, we analyze the stability of the base crop sizes. We shift the intercept values $\beta_1$ and $\beta_2$ by $\pm 0.2$ from the default settings to simulate tighter or looser base crops.

Table~\ref{tab:sensitivity_beta} illustrates the robustness of Visual Funnel. Tighter crops ($\beta - 0.2$) lead to a slight performance drop due to the severance of immediate local context. However, the performance variance across the tested range is minimal ($<0.6\%$), indicating that our method does not require precise hyperparameter tuning to achieve significant gains.

\begin{table}[h]
    \centering
    \small
    \setlength{\tabcolsep}{8pt}
    \resizebox{\linewidth}{!}{
    \begin{tabular}{lcccc}
        \toprule
        Base Scale Shift & $\beta_1$ & $\beta_2$ & DocVQA Acc. (\%) & $\Delta$ \\
        \midrule
        Tighter Crops ($-0.2$) & 1.0 & 1.4 & 60.5 & -0.6 \\
        \textbf{Default} & \textbf{1.2} & \textbf{1.6} & \textbf{61.1} & \textbf{--} \\
        Wider Crops ($+0.2$) & 1.4 & 1.8 & 60.9 & -0.2 \\
        \bottomrule
    \end{tabular}}
    \caption{\textbf{Robustness of Base Expansion Factors.} Shifting the base crop size $\beta$ shows minimal impact on performance, demonstrating the method's stability.}
    \label{tab:sensitivity_beta}
\end{table}

\begin{table*}[t]
    \centering
    \small
    % -------------------------------------------------------
    % [Table 5] - 위쪽 표
    % -------------------------------------------------------
    % 표 너비를 전체(linewidth)로 설정. 너무 넓어 보이면 0.8\linewidth 등으로 조절 가능
    \resizebox{0.7\linewidth}{!}{ 
        \setlength{\tabcolsep}{4pt} 
        \begin{tabular}{clccc}
            \toprule
            \# Crops ($K$) & Configuration & Token Usage & DocVQA Acc. & $\Delta$ \\
            \midrule
            0 & Original Image Only & $1\times$ & 51.5 & -9.6 \\
            1 & Focal Only & $\sim 1.3\times$ & 55.1 & -6.0 \\
            2 & Focal + Imm. & $\sim 1.6\times$ & 58.0 & -3.1 \\
            \textbf{3} & \textbf{Focal + Imm. + Broader} & $\mathbf{\sim 1.9\times}$ & \textbf{61.1} & \textbf{--} \\
            4 & + Global Context & $\sim 2.2\times$ & 60.7 & -0.4 \\
            \bottomrule
        \end{tabular}
    }
    \caption{\textbf{Impact of Portfolio Size ($K$).} Increasing crops saturates at $K=3$. Adding more leads to a ``Redundancy Penalty.''}
    \label{tab:portfolio_size}

    \vspace{15pt} % 두 표 사이의 간격 (취향에 맞게 조절: 10pt ~ 20pt 추천)

    % -------------------------------------------------------
    % [Table 6] - 아래쪽 표
    % -------------------------------------------------------
    \resizebox{\linewidth}{!}{
        \begin{tabular}{lccccc}
            \toprule
            Model Configuration & Avg. Tokens & Latency (ms) & Relative Time & DocVQA Acc. & \textbf{Gain/Time} \\
            \midrule
            Base (No Crop)      & $\sim 1,200$ & 450  & $1.00\times$ & 51.5 & -- \\
            w/ ViCrop           & $\sim 1,800$ & 780  & $1.73\times$ & 54.2 & Low \\
            w/ ViCrop (Top-3)   & $\sim 2,400$ & 920  & $2.04\times$ & 55.3 & Low \\
            \midrule
            \textbf{w/ Visual Funnel (Ours)} & $\sim 2,300$ & \textbf{890} & \textbf{1.98$\times$} & \textbf{61.1} & \textbf{High} \\
            \bottomrule
        \end{tabular}
    }
    \caption{\textbf{Efficiency vs. Performance Trade-off.} Compared to the Base model, Visual Funnel requires approximately $2\times$ the inference time but yields a massive performance gain (+9.6\%). Notably, it is more efficient than the naive multi-crop baseline (ViCrop Top-3) in terms of accuracy per computational unit.}
    \label{tab:efficiency_latency}
\end{table*}

\section{Ablation on Portfolio Size}
\label{sec:appendix_portfolio_size}

In Visual Funnel, we construct a hierarchical portfolio consisting of three specific crops: \textit{Focal} ($\mu_0$), \textit{Immediate Context} ($\mu_1$), and \textit{Broader Context} ($\mu_2$), in addition to the original image. A critical question arises: \textit{Is the performance gain simply due to the increased quantity of visual tokens, or is the three-layer hierarchical structure optimal?}

To answer this, we evaluate the impact of the number of portfolio crops ($K$) on the DocVQA dataset using Qwen2.5-VL-3B-Instruct. We incrementally add crops following our hierarchical expansion strategy:
\begin{itemize}
    \item $K=1$: Focal crop only (similar to standard ViCrop).
    \item $K=2$: Focal + Immediate Context.
    \item $K=3$: Focal + Immediate + Broader Context (\textbf{Ours}).
    \item $K=4$: Focal + Immediate + Broader + Global Context (an even wider crop).
\end{itemize}

% \begin{table}[t]
%     \centering
%     \small
%     \resizebox{\linewidth}{!}{
%         \setlength{\tabcolsep}{4pt} % 컬럼 간격도 조금 줄임
%         \begin{tabular}{clccc}
%             \toprule
%             \# Crops ($K$) & Configuration & Token Usage & DocVQA Acc. & $\Delta$ \\
%             \midrule
%             0 & Original Image Only & $1\times$ & 51.5 & -9.6 \\
%             1 & Focal Only & $\sim 1.3\times$ & 54.2 & -6.9 \\
%             2 & Focal + Imm. & $\sim 1.6\times$ & 58.0 & -3.1 \\
%             \textbf{3} & \textbf{Focal + Imm. + Broader} & $\mathbf{\sim 1.9\times}$ & \textbf{61.1} & \textbf{--} \\
%             4 & + Global Context & $\sim 2.2\times$ & 60.7 & -0.4 \\
%             \bottomrule
%         \end{tabular}
%     }
%     \caption{\textbf{Impact of Portfolio Size ($K$).} Increasing crops saturates at $K=3$. Adding more leads to a ``Redundancy Penalty.''}
%     \label{tab:portfolio_size}
% \end{table}

As presented in Table~\ref{tab:portfolio_size}, the results support our structural design:

\begin{enumerate}
    \item \textbf{Significant Gain from Hierarchy ($K=1 \rightarrow 3$):} Moving from a single focal crop ($K=1$) to our three-layer portfolio ($K=3$) yields a substantial improvement (+6.0\%). This confirms that resolving Contextual Blindness requires not just the high-resolution detail of the target, but also the intermediate scales that bridge the detail to the global view.
    
    \item \textbf{The Redundancy Penalty ($K=4$):} Interestingly, adding a fourth crop ($K=4$) does not further improve performance; in fact, it leads to a slight degradation (61.1\% $\rightarrow$ 60.7\%). We attribute this to the \textit{Redundancy Penalty}: providing too much overlapping visual information can overwhelm the MLLM's attention mechanism, causing it to lose focus on the critical details.
    
    \item \textbf{Efficiency Trade-off:} Furthermore, $K=4$ increases the input token count and inference latency without functional benefit. Therefore, we identify $K=3$ as the optimal configuration that maximizes structural diversity while maintaining computational efficiency.
\end{enumerate}

\section{Computational Efficiency Analysis}
\label{sec:appendix_efficiency}

While Visual Funnel significantly enhances fine-grained perception, it inevitably introduces computational overhead due to the two-step inference process and the processing of additional visual tokens. In this section, we provide a detailed analysis of inference latency and token usage to demonstrate the cost-effectiveness of our approach.

\textbf{Experimental Setup.} We measured the average wall-clock time per query on the DocVQA validation set. All experiments were conducted on four NVIDIA RTX PRO 6000 (96GB) GPU with PyTorch 2.8. The latency includes image preprocessing, visual encoding, and language generation.

% \begin{table}[h]
%     \centering
%     \small
%     \resizebox{\linewidth}{!}{
%         \begin{tabular}{lccccc}
%             \toprule
%             Model Configuration & Avg. Tokens & Latency (ms) & Relative Time & DocVQA Acc. & \textbf{Gain/Time} \\
%             \midrule
%             Base (No Crop)      & $\sim 1,200$ & 450  & $1.00\times$ & 51.5 & -- \\
%             w/ ViCrop           & $\sim 1,800$ & 780  & $1.73\times$ & 54.2 & Low \\
%             w/ ViCrop (Top-3)   & $\sim 2,400$ & 920  & $2.04\times$ & 55.3 & Low \\
%             \midrule
%             \textbf{w/ Visual Funnel (Ours)} & $\sim 2,300$ & \textbf{890} & \textbf{1.98$\times$} & \textbf{61.1} & \textbf{High} \\
%             \bottomrule
%         \end{tabular}
%     }
%     \caption{\textbf{Efficiency vs. Performance Trade-off.} Compared to the Base model, Visual Funnel requires approximately $2\times$ the inference time but yields a massive performance gain (+9.6\%). Notably, it is more efficient than the naive multi-crop baseline (ViCrop Top-3) in terms of accuracy per computational unit.}
%     \label{tab:efficiency_latency}
% \end{table}

\textbf{Analysis.} As shown in Table~\ref{tab:efficiency_latency}:

\begin{itemize}
    \item \textbf{Latency Overhead:} Visual Funnel increases the inference latency by approximately $1.98\times$ compared to the base model. This is primarily due to the additional forward pass required for \textit{Contextual Anchoring} (Step 1) and the encoding of the multi-scale portfolio (Step 2).
    
    \item \textbf{Comparison with Baselines:} Compared to \texttt{w/ ViCrop (Top-3)}, which processes a similar number of visual tokens, our method is slightly faster ($890$ms vs. $920$ms) and significantly more accurate ($61.1\%$ vs. $55.3\%$). This indicates that the \textit{structure} of the visual input is more important than raw pixel quantity.
    
    \item \textbf{Parallelization:} It is worth noting that the multiple crops in Step 2 are encoded in a single batch, allowing us to leverage GPU parallelism. This ensures that the latency does not scale linearly with the number of crops.
    
    \item \textbf{Practicality:} Given the complexity of fine-grained tasks (e.g., reading small text in documents), we argue that a $2\times$ latency increase is a justifiable trade-off for a $\sim10\%$ accuracy improvement. For real-time applications, Visual Funnel can be selectively applied only when the base model's confidence is low.
\end{itemize}

% \begin{figure*}[!t]
%   \centering
%      \includegraphics[width=1.0\linewidth]{supples/appendix_figure.pdf}
%     \caption{Success (first 6) and failure (last 3) examples of Qwen2.5-VL-3B-Instruct.}
%     \label{fig:qualitatives examples}
% \end{figure*}

\begin{figure*}[t]
    \centering
    \includegraphics[width=\linewidth]{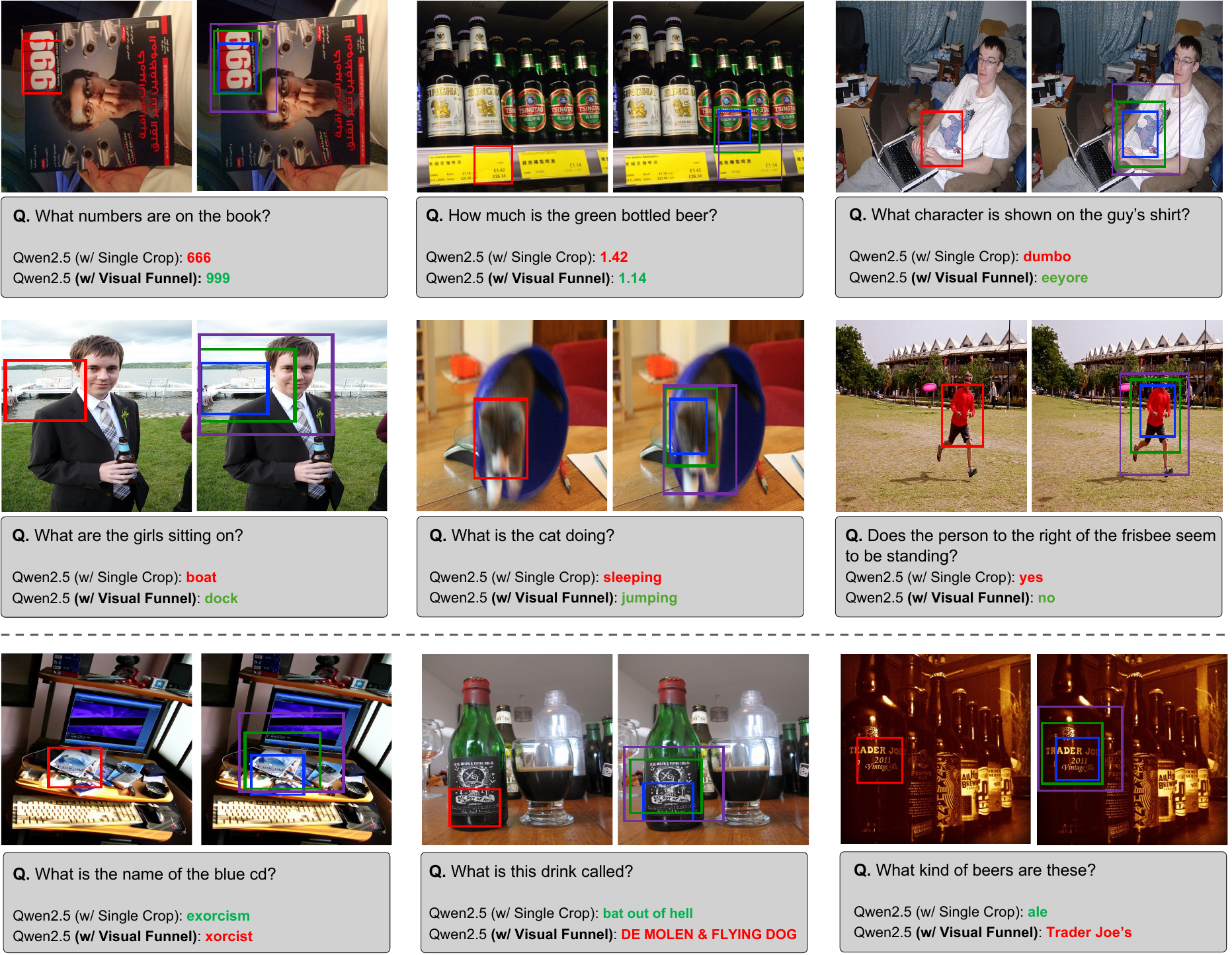} % 파일명은 실제 파일명으로 수정하세요
    \caption{\textbf{Qualitative comparison between the Single-Crop baseline and Visual Funnel.} 
    We visualize the inputs and predictions using \texttt{Qwen2.5-VL-3B-Instruct}. The \textcolor{red}{Red box} represents the input for the standard Single-Crop baseline (w/ ViCrop), while the \textcolor{blue}{Blue}, \textcolor{green}{Green}, and \textcolor{violet}{Purple} boxes represent the hierarchical portfolio (Focal, Immediate, Broader context) used in Visual Funnel.
    \textbf{(Top two rows) Success Cases:} Visual Funnel successfully resolves \textit{Contextual Blindness} across various tasks, including fine-grained OCR (e.g., identifying ``999'' instead of inverted ``666''), small object recognition (``Eeyore''), and action/state reasoning (``jumping'' vs. ``sleeping'', ``dock'' vs. ``boat'').
    \textbf{(Bottom row) Failure \& Ambiguous Cases:} Examples below the dashed line illustrate limitations where the model still struggles despite improved context. These include partial OCR errors (``xorcist''), ambiguity in label hierarchy (Brewery name vs. Drink name), or distinct object attributes (Brand vs. Type), suggesting directions for future work.}
    \label{fig:qualitatives examples}
\end{figure*}

\section{Qualitative Visualizations}
\label{sec:appendix_efficiency}
We present further qualitative success and failure cases of \texttt{Qwen2.5-VL-3B-Instruct} in Figure~\ref{fig:qualitatives examples}.

\end{document}